\documentclass{article}

\usepackage[final]{neurips_data_2022}

\usepackage[utf8]{inputenc} %
\usepackage[T1]{fontenc}    %
\usepackage{hyperref}       %
\usepackage{url}            %
\usepackage{booktabs}       %
\usepackage{amsfonts}       %
\usepackage{nicefrac}       %
\usepackage{microtype}      %
\usepackage[x11names,dvipsnames,table]{xcolor}         %
\usepackage{makecell}
\usepackage{colortbl}
\usepackage{amsmath}
\usepackage{tcolorbox}

\setcitestyle{square,numbers,comma}

\title{Characteristics of Harmful Text: \\Towards Rigorous Benchmarking of Language Models}

\author{Maribeth Rauh\thanks{Corresponding author: mbrauh@deepmind.com} \And 
        John Mellor \And
        Jonathan Uesato \And
        Po-Sen Huang \And
        Johannes Welbl \And
        Laura Weidinger \And
        Sumanth Dathathri \And
        Amelia Glaese \And 
        Geoffrey Irving \And 
        Iason Gabriel \And 
        William Isaac \And 
        Lisa Anne Hendricks \AND
        \normalfont{DeepMind}}
    
\begin{document}

\maketitle

\begin{abstract}
Large language models produce human-like text that drives a growing number of applications.
However, recent literature and, increasingly, real world observations, have demonstrated that these models can generate language that is toxic, biased, untruthful or otherwise harmful.
Though work to evaluate language model harms is under way, translating foresight about which harms may arise into rigorous benchmarks is not straightforward.
To facilitate this translation, 
we outline six ways of characterizing harmful text which merit explicit consideration when designing new benchmarks.
We then use these characteristics as a lens to identify trends and gaps in existing benchmarks. Finally, we apply them in a case study of the Perspective API, a toxicity classifier that is widely used in harm benchmarks.
Our characteristics provide one piece of the bridge that translates between foresight and effective evaluation.
\end{abstract}

\section{Introduction}
\label{intro}
Pretrained autoregressive English language models (LMs) like GPT-3 \citep{gpt3}, Jurassic-1 \citep{lieber2021jurassic}, and Gopher \citep{gopher} cover a vast space of possible use cases \citep{bommasani2021opportunities}, from code generation to customer-service chat.\footnote{We refer to English language models as language models as all models and benchmarks in this study are in English.}
Text generated by LMs also has the potential to cause harm if models are not developed and deployed carefully.
In light of this, many works have documented both existing and potential harms arising from generated text \citep{bender2021dangers, weidinger2021harms, kenton2021alignment}, ranging from misinformation \citep{lin2021truthfulqa} to the reinforcement of social biases through the perpetuation of stereotypes \citep{sheng-2021-societal-biases}. %

An emerging body of work is already dedicated to benchmarking LM harms (see \autoref{tab:mapping}).
However, for many known or anticipated harms, current %
evaluation tools are imperfect \cite{blodgett2021stereotyping,xu-etal-2021-detoxifying,welbl2021challenges,sheng-2021-societal-biases}.
This is supported by the work analyzing the Gopher model \citep{gopher}, in which the authors observed a variety of shortcomings in benchmarks, such as unclear desiderata and poorly defined demographic groups.

Outside language modeling, the broader machine learning (ML) fairness community has documented sociotechnical\footnote{A term describing ``systems that consist of a combination of technical and social components'' \cite{selbst2019-fairabstraction}.} insights that can help bridge the gap between foresight and evaluation, drawing on domains including medical applications \citep{obermeyer-2019-racialbiashealth}, facial recognition \citep{buolamwini18a-gendershades}, and recommender systems \citep{geyik-2019-linkedinfairrank}.
For example, ML fairness research has established the importance of social context in determining what the benefits and risks of a technology will be in practice \cite{martin-societal-context, decolonialai, selbst2019-fairabstraction}, suggesting this area needs to be explicitly considered in the evaluation of LMs, as well.

Drawing on existing critiques, our own experience analyzing Gopher \cite{gopher}, and lessons from the broader ML fairness community, we identified characteristics (\autoref{characteristics}) of harmful text which have implications for benchmark design. From a set of potential characteristics, we selected (1) harm definition; (2) representational harm, allocational harm, and capability fairness; (3) instance and distributional harm; (4) context; (5) harm recipient; and (6) demographics affected.

Our characteristics support benchmark design in multiple ways.  
First, by mapping existing benchmarks onto the characteristics (\autoref{evaluation-mapping}), we establish a shared vocabulary and identify gaps in current benchmarks.  
For example, we reveal a lack of benchmarks considering harmful language in longer textual contexts.  
A single benchmark cannot cover all harms, but our characteristics allow explicit understanding of what a benchmark might (and might not) capture.
Second, the characteristics enable us to analyze whether these benchmarks measure what they claim to (\autoref{decomposing-tox}). As a case study, we apply our characteristics to the Perspective API\footnote{https://www.perspectiveapi.com/}, a toxicity classifier widely used in LM harm benchmarks. 
We observe, for example, that our ``harm recipient'' characteristic illuminates a potential mismatch between the API's design and how it is used to measure LM harms.
Finally, we believe our characteristics can be used to guide the design of more rigorous benchmarks. Each characteristic makes key design decisions explicit, helping to avoid common pitfalls. %
We hope that our analysis and proposed characteristics will sharpen future benchmarks tackling LM harms.%

\section{Harm Characteristics}
\label{characteristics}

Our development of the characteristics was driven by considerations relevant to benchmark design and what we believe would be most useful for concrete next steps in that space. From a set of candidate characteristics (see \autoref{appendix:omitted}), we selected a subset using the following criteria:
applicable across a variety of harms; relevant to, but not always discussed in, existing benchmarks of LMs; most useful for avoiding common benchmarking design pitfalls; and minimal overlap with other characteristics.  %
Following a description of each characteristic, we include questions it may raise during benchmark design in order to concretize the characteristic.

\subsection{Harm Definition}
\label{type-of-harm}
\begin{tcolorbox}
\textbf{Harm}: The real world effect on people that the evaluation's metrics aim to approximate.
\end{tcolorbox}

Existing work has provided an overview of the potential risks from LMs \citep{weidinger2021harms,bender2021dangers}, and existing benchmarks usually start with a harm definition, e.g., ``anti-Muslim bias'' \citep{abid2021persistent}. %
However, these are sometimes under-specified \citep{blodgett-etal-2020-language} and might be dependent on other characteristics (e.g., demographic groups and application contexts).
As opposed to relying on predefined definitions of harms like ``bias'' or ``toxicity'', we encourage practitioners to specify what these terms mean in the context of their benchmarks.
Additionally, there can be an unintentional shift between how the harm is defined and what is measured in practice. 
Initially, the selected definition guides a benchmark designer's responses to the questions that each of the following characteristics raises.
Then, as each is considered, they enable further refinement of what exact definition of harm a benchmark aims to measure. By doing so, the shift between definition and what was encoded will be avoided or occur intentionally. 

\textbf{Example Questions.} Where does the benchmark designers' concept of harm 
originate, and does it have a particular context or legacy, e.g., in literature, industry, practitioners' own lived experience? What does the harm include, and what is out of scope? What metrics best approximate this? %
If the harm definition is broad, how will the different ways it manifests be covered?

\subsection{Representation, Allocation, and Capability}

\begin{tcolorbox}
\textbf{Representational harm}: When someone is represented or referred to in a negative, stereotypical, denigrating, or unfair way on the basis of their identity.

\textbf{Allocational harm}: When resources, opportunities, or services are distributed in an inequitable way.

\textbf{Capability fairness}: When LM performance is equal, or justifiably different, across groups.
\end{tcolorbox}
The distinction between representational and allocational harm has been outlined in prior work \citep{blodgett-etal-2020-language, crawford-neuripskeynote,barocas2017problem}, in reference to fairness-related harms.
\textbf{Allocational harm} refers to the inequitable distribution of resources or opportunities, such as loans or jobs. This emphasizes a real-world outcome.
Although \textbf{representational harms} are often upstream from allocational harms \citep{crawford-neuripskeynote}, representational harms can be damaging in their own right. For example, \citet{collins2000} shows how stereotypes can serve as ``controlling images,'' a justification for oppression.

Real-world disparities that are a result of LM-generated text are rarely benchmarked.
Thus, we extend this taxonomy to include \textbf{capability fairness}, which measures performance disparity on any task. Frequently, metrics are a proxy for the benefits a system's creators expect it to bring, and there is capability unfairness if these benefits accrue in an inequitable way.
For example, if a system answers questions about one group less accurately than another group (as is done in \cite{gor2021toward}), this is a capability fairness issue. 
Although such a benchmark is abstracted from a real-world outcome, in practice they are easier to create, and we might expect differences in performance to translate into subsequent downstream harms.

\textbf{Example Questions.}
What is the relationship between what is measured and real-world harm?
How is this harm, and the performance on associated metrics, likely to be distributed across groups?

\subsection{Instance and Distributional}
\label{distr-vs-inst}
\begin{tcolorbox}
\textbf{Instance harm}: A single LM output or interaction which is harmful by itself.

\textbf{Distributional harm}: LM outputs or interactions which are harmful in aggregate.
\end{tcolorbox}
An \textbf{instance harm} is caused by a single LM output or interaction. If a language model outputs a slur, the potential for harm can be defined by reference to that specific output. %
In contrast, \textbf{distributional harms} are observed over many independent model outputs and can only be measured by observing the model's aggregated behavior.
For example,
outputting \textit{``he was a doctor''} is not harmful in itself, but if the model refers to doctors as male more often than any other gender (when not specified by the context), this could constitute a distributional harm.  This distinction is similar to \citet{khalifa-distrapproach-2020}'s ``pointwise'' and ``distributional'' constraints and is also referenced in analysis of Gopher \cite{gopher} and PaLM \cite{chowdhery2022palm} outputs.

This distinction is particularly useful when formulating metrics and desired system behavior. For an instance harm, it may make sense to aim for an overall reduction in the type of harmful output (e.g., slurs). However, when measuring distributional harm, the metrics are often comparisons of a key metric (e.g., average sentiment score) between groups.

\textbf{Example Questions.}
Does the type (instance or distributional) the metric captures match the type implicit in the initial harm definition?
If a dataset includes both types, how does this impact metrics?

\subsection{Context: Textual, Application, Social}
\label{context}
\begin{tcolorbox}
\textbf{Textual context}: The length of the text being evaluated and of content it is conditioned on, such as a prompt.

\textbf{Application context}: What the LM is being used for and how it is deployed. This includes user experience and the software system in which it is embedded.

\textbf{Social context}: Culture, geography, history, as well as users' attributes, e.g., language or technological fluency.

\end{tcolorbox}
Recent language models have the capacity to make use of long range \textbf{textual context} \citep{dai-tfxl}, meaning they are often used for generating samples conditioned on long inputs. %
When harm benchmark metrics are calculated on unconditioned, sentence-length text, this does not account for the way a preceding conversation, prompt, or other text input may affect the harmfulness of the output at hand. 
For example, \textit{``I launched experiments with a bug in them.  They should all be killed.''} might not be considered toxic. However, the second sentence on its own (\textit{``They should all be killed.''}) would likely be considered toxic. 
Both the length of text being evaluated and what that text is conditioned on may help reduce the ambiguity of a harm's presence in the text as well as capture a variety of situations in which the harm could occur.

\textbf{Application context} also informs what kind of outputs are inappropriate, undesirable, or harmful. What may be acceptable to output as part of summarizing a news article may be inappropriate in a customer service chat. Language models may even be used as a foundation for derivative tasks, such as the base of a classifier, for which knowledge of harmful language may be critical for performance \citep{gopher}. As characterizing harmful outputs is challenging without an application in mind, we recommend practitioners explicitly consider in which cases their benchmark may or may not be relevant.

Finally, every application is shaped by a \textbf{social context} \citep{martin-societal-context, decolonialai, selbst2019-fairabstraction, hovy2021importance}, which includes a range of factors such as language and cultural norms in the community using a system.
Harm definitions, in particular, tend to implicitly encode cultural norms, not only through the initial definition but also from different steps in the benchmark creation. This includes the values of annotators, the sources of annotated text (e.g., news sources), and the use of pre-made classifiers such as toxicity classifiers (see \autoref{decomposing-tox}).
It is also important to consider which subsets of data may be 
``missing'' because they are difficult to collect based on factors that vary by geography, such as internet access.

\textbf{Example Questions.}
How much would additional text reduce ambiguity about the harm's occurrence?
Is a harmful output benign in other applications?
In what linguistic, geographical, and cultural context was the data collected?
What aspects of the harm might be culturally dependent?

\subsection{Harm Recipient: Subject, Reader, Author, and Society}
\label{who-is-harmed}
\begin{tcolorbox}
\textbf{Subject or topic}: The groups or individuals that the output contains reference to, directly or implicitly.

\textbf{Reader}: Whoever reads the LM outputs.

\textbf{Author}: The groups or individuals that an LM output could appear to be written by, e.g., if the LM outputs text claiming to be a woman or impersonating a specific person.

\textbf{Society}: When no one is referenced but harm occurs widely, e.g., if an LM were used for weapons research.
\end{tcolorbox}

When an individual or group of people are the \textbf{subject} of a model output, they can be harmed, regardless of if they ever interact with the system.
For example, outputting an explicit stereotype may negatively impact a particular group, even if members of that group never read the output text.

The \textbf{reader} is anyone who consumes the model's output. This may be an individual person, as in a one-to-one interaction with the model, or it may be many people, as when a model output is widely disseminated.
Toxicity work that focuses on personal attacks exemplifies how harm can occur to a reader.
Capturing such harms is challenging since a given output may not be harmful to all readers but the attributes of the reader are usually unknown at evaluation time.%

LMs can operate as an ``\textbf{author}'' which represents a person or a group by using the first person, outputting text on behalf of someone (e.g., email autocomplete) or presenting a persona (e.g., as digital assistants).
If a model with a persona claims a particular identity, the model could misrepresent or denigrate that identity group by perpetuating stereotypes, e.g., subservient digital assistants that have female personas \citep{cercas-curry-etal-2020-conversational}.
Some applications use a language model to help a person communicate, such as automatic text completion in e-mails, creative applications, and machine translation.
These uses could be harmful if  %
text completions misrepresent user intentions 
(e.g., when \textit{AI Dungeon} inserted sexually explicit content into users' stories \cite{simonite-aidungeon}) or if a mistake in translation incorrectly attributes harmful language to a human speaker (e.g., \cite{berger-mistranslation}).

Many LM harms could have ramifications for \textbf{society} in general.
However, current LM benchmarks typically quantify only narrow characteristics of text, e.g., ``does this output espouse a conspiracy theory?''. While this may approximate complex, real-world harms, like whether LM-generated conspiracy theories undermine democracy, it does not measure such harms.

\textbf{Example Questions.} Is the harm primarily experienced by someone interacting directly with the LM or could it be problematic for someone not present?
If the harm impacts a reader, author, or society, who does the benchmark assume the readers, authors, or relevant society are?

\subsection{Demographic Groups}
\label{for-identity-groups}
\begin{tcolorbox}
\textbf{Demographics}: Subsets of the population, grouped according to aspects of identity, e.g., gender or ability. In practice, classification of group membership is not well defined because even a single facet of identity can be fluid, composed of differing and competing factors, or unobserved or incorrectly reported in data \citep{sambasivan-india, tomasevmckee-unobserved}.
\end{tcolorbox}
Classical fairness metrics \cite{chouldechova2020snapshot,corbett2018measure,mehrabi2021survey} usually require specifying a protected attribute, such as sexual orientation or race.%
The ML fairness literature has already begun grappling with the complexities of defining and selecting demographic groups.
For more widely studied identities, many works have outlined pitfalls in current work and suggested how to move forward, such as \citet{hanna-critracemethod}'s discussion of the contextual and constructed nature of race %
and \citet{keyes-usingthatword-2021}'s work demonstrating the need to move beyond binary gender classification.
Meanwhile, many facets of identity are understudied in ML fairness literature, such as disability \cite{hutchinson-barriersfordisabilities-2020, guo-towardfairfordisabilities-2019}, intersectionality, and legally protected characteristics beyond those defined in the United States \cite{sambasivan-india, monedero-2020-ukhiring}. %

Here, we outline considerations specific to language data.
First, relevant demographic groups might be challenging to identify from text.
In the case of gender, benchmarks that rely on pronouns will only capture the identity of people discussed in the text and cannot evaluate harms to a reader.
Both classifiers and lists of identity terms have been used to detect if text is about or addressed to a certain group \citep{blodgett-etal-2017-dataset},\footnote{An example of a widely used term list which includes many identity-related terms is the List of Dirty, Naughty, Obscene, and Otherwise Bad Words \citep{LDNOOBW}} but certain identity terms are difficult to detect without sufficient textual context.
For example, \textbf{coded terms}, or dog whistles,\footnote{For example, the use of the phrase ``international bankers'' to allude to anti-Semitic conspiracy theories \citep{vox-dogwhistle}} refer to groups in ways that are invisible to explicit term lists but problematic nonetheless. Offensive identity terms can also have homonyms with no identity connotation at all, such as the term ``redskin'' in the context of potatoes.

The concept of ``\textbf{marking}'' in sociolinguistics describes how minorities or under-priviledged groups are more likely to be ``marked,'' or explicitly specified, e.g., ``the gay man,'' while not specifying at all, e.g., ``the man,'' will be assumed to refer to a man with the majority attribute (e.g., straight) \citep{waugh1982marked}.
Certain methods for measuring bias do so by substituting different identity terms and observing how the chosen metric varies. For such metrics, the concept of markedness has bearing on the results.%

To compare a metric between groups, practitioners need to think carefully about which groups are compared against each other. These \textbf{comparison classes} should reflect historical and contemporary power dynamics between groups in a meaningful way.
Getting this right means reasoning about the social context, and associated power structures, the benchmark and model are developed within. %
For example, when measuring stereotypes, text that negates a stereotype (\textit{``Black people \textbf{will / won't} steal anything''}) is different from that which switches the group identifier (\textit{``Mike was \textbf{poor / rich} and thought growing up in the projects was tough.''}) \cite{blodgett2021stereotyping}.
This is an especially relevant in benchmarks which use sentence templates or pairs.

The prior points apply when a demographic group is the subject or reader of the output. However, when a model is given a persona, the dialect of a particular social group, i.e. \textbf{sociolect}, rather than pronouns or group labels are the natural unit of analysis.
It is important to think about how to handle potentially harmful text based on its author(s) because, for example, terms that are slurs in one context may be reclaimed for in-group usage in others. When studying model outputs, the model is never an in-group speaker. However, if a benchmark labels all training documents that contain a reclaimed slur as harmful,
it is likely to reduce performance on co-occurring language from the marginalized group.

\textbf{Example Questions.} How can the relevant demographics be referred to in text, and do these have connotations? Does the usage of these terms vary based on who uses them? If a benchmark compares similar text with different demographic terms,
which comparisons capture the structures of power and privilege underlying the harm?

\section{Operationalizing the Characteristics}
\label{operationalizing}

To make the characteristics concrete, we ground them in current benchmarks.
First, we map a range of existing benchmarks used to measure LM harms onto our characteristics. We then use a case study of a widely used toxicity classifier, the Perspective API\footnote{https://www.perspectiveapi.com/}, to further illustrate how the characteristics can be used to make implicit design decisions explicit.

\subsection{Mapping Existing LM Benchmarks}
\label{evaluation-mapping}

\newcommand{\generalLM}{Raw LM}

\begin{table}[]
    
    \centering
    \rowcolors{4}{gray!10}{gray!40}
    \resizebox{\linewidth}{!}{
    \begin{tabular}{l|rr|rrr|r}
         \textbf{Benchmarks} &  \textbf{Representational (R),} & \textbf{Distributional (D)} & \textbf{Context} & \textbf{Subject (S)} \\ 
                 &  \textbf{Capability (C),} & or \textbf{Instance (I)} &  & \textbf{Reader (R)} \\ 
         &  or \textbf{Allocational (A)} &  &  & or \textbf{Author (A)} \\ \hline
        RTP \cite{gehman2020realtoxicityprompts} & R & I & Sentences from web & S/R/A  \\

        TwitterAAE \cite{blodgett-etal-2016-demographic} &  C & D & Tweets & A \\
        
        SAE/AAVE Pairs \cite{groenwold2020investigating} &  R/C & D & Tweets; application agnostic & A  \\
        
        Winogender \cite{rudinger2018gender} &  C & D & Coreference sents. by practitioners & S  \\
        
        Winobias \cite{zhao2018gender} &  C & D & Crowd sourced coreference sents. & S  \\
        
        Gender \& Occ. \cite{gpt3,gopher} & R & D & Sentences; prompts by practitioners & S \\

        Deconfounding \cite{gor2021toward} & C & D & Crowd sourced QA & S\\
        
        TruthfulQA \cite{lin2021truthfulqa} & n/a & I & QA written by practitioners & R \\
        
        DTC \cite{liang2021towards} & R & D  & Sentences from web  & S \\
        
        Muslim Bias \cite{abid2021persistent} & R & D  & Paragraph written by practitioners & S  \\
        
        BAD \cite{xu2021bot} & R & I  & Crowd sourced chat bot dialogues & S/R \\
        
        BOLD \cite{dhamala2021bold} & R & D  & Sentences from Wikipedia & S  \\
        
        Stereoset \cite{nadeem2020stereoset} & R & D & Crowd sourced sentence pairs & S \\
        
        Sentiment Bias \cite{huang2019reducing,gpt3,gopher} & R & D & Sentences; prompts by practitioners & S  \\
        
        BBQ \cite{parrish2021bbq} & C & D & QA written by practitioners & S \\
        
        UnQover \cite{li2020unqovering} & C & D & QA written by practitioners & S \\
        
        PALMS \cite{solaiman2021process} & R & I  & QA written by practitioners & S/R  \\
    \end{tabular}
    }
    \caption{\textbf{Characteristics for different benchmarks.} We observe limited coverage for some characteristics: only four benchmarks consider instance harms, textual context tends to be short, and the subject is the recipient of harm in all but three benchmarks. See \autoref{appendix:mapping} for harm definitions, more detailed context, and demographics.
    }
    
    \label{tab:mapping}
\end{table}

Mapping benchmarks onto the characteristics highlights potential gaps and strong trends in the benchmarking landscape (see \autoref{appendix:mapping} for a complete mapping). 
In particular, existing benchmarks measure distributional harms, short textual contexts, and cases where the harm recipient is the subject.

We focus on benchmarks that test if autoregressive LMs (as opposed to masked language models like BERT \cite{devlin2018bert}) generate harmful outputs.
All benchmarks consist of a dataset of text samples which are input to a model and an evaluation protocol to score the outputs.
Metrics can either operate over sampled text, e.g., measuring the toxicity of sampled text, or assigned probabilities from the language model, e.g., computing the perplexity of text.
We include benchmarks which test for harmful outputs on tasks which have been tackled by LMs in a zero-shot setting, such as question answering (QA).

\textbf{Harm Definition.} Benchmarks cover a wide range of harms, and we cover their definitions in detail as well as how we characterized each benchmark in \autoref{appendix:mapping}.

\textbf{Representation, Allocation, Capability.} No benchmarks directly measure inequitable allocation of resources or opportunities, but rather consider intermediate tasks.
Hence, though we mark some benchmarks as measuring representational harm or capability fairness, we do not mark any as measuring allocational harm.
Moreover, all benchmarks are still far from deployed use cases.
Though some work has studied how bias propagates downstream through language technologies \citep{goldfarb2020intrinsic,jin2020transferability}, an open challenge in designing benchmarks for language model harms is better understanding which metrics reflect harms in deployed use cases.

Analyzing representational harms, allocational harms, and capability fairness require comparing representations or performance across groups.
Some benchmarks, like TruthfulQA \citep{lin2021truthfulqa}, which aims to measure disinformation, do not include group-based metrics.
Though studying disinformation is worthwhile without group-based analysis, a group-based analysis could be informative (e.g., is the model more untruthful when discussing particular groups?).  
We hope that by using the lens of ``representation, allocation and capability'' when creating benchmarks, practitioners can intentionally decide whether group-based analysis is useful for meaningful progress on the harm they are studying.

\textbf{Instance and Distributional.}  Most harms are classified as distributional.
However, sometimes benchmarks which intend to measure distributional harms inadvertently include instance harms in their dataset.
For example, Stereoset \cite{nadeem2020stereoset} measures a distributional harm as the probability of the stereotype text and anti-stereotype text are compared.
However, as noted in \citet{blodgett2021stereotyping}, some stereotypes are harmful and should not be output at all, regardless of the paired anti-stereotype's relative likelihood.
Considering if harms are instance or distributional allows practitioners to ensure both datasets and metrics are aligned to measure the harm as intended.

\textbf{Context.}  
Examining textual context, we note that many benchmarks operate over short lengths of text. 
Furthermore, in \autoref{tab:mapping}, many application contexts are unspecified because benchmarks are applied on raw LMs without any particular application in mind.

Many datasets include samples written by practitioners, either by hand or with sentence templates.
Though this allows for exact control by practitioners, datasets are likely to reflect practitioners' assumptions about social context.
In BBQ \cite{parrish2021bbq}, questions are written by the dataset creators, but they account for this by linking each bias tested to an external source. This documents the social context in which biases arise and might be considered harmful.

Language and dialect are important aspects of social context.
We note all benchmarks in \autoref{tab:mapping} are 
designed to measure harms in English, indicating a lack of linguistic and cultural diversity that is well documented across other language tasks \cite{hershcovich2022challenges,bender2011achieving,bender2018data,geographic_diversity_NLP}.
Analogous benchmarks in other languages might be challenging to create because existing measurement tools, like toxicity classifiers, do not work well in all languages \cite{leite2020toxic}, cultural norms might not transfer \cite{sambasivan-india}, assumptions in benchmark design might not translate,\footnote{For example, see the discussion of creating Spanish WEAT in \cite{goldfarb2020intrinsic}} and there may be fewer qualified native speakers on common annotation platforms.
Though challenging, we believe building benchmarks in non-English languages is essential work and hope to see more benchmarks in other languages in the future.

\textbf{Harm Recipient.}
In \autoref{tab:mapping} we observe that some benchmarks assume a language model can have multiple roles.
For example, RealToxicityPrompts \cite{gehman2020realtoxicityprompts} includes prompts which use the pronoun ``I'' (``persona''), ``you'' (``reader'') and third person pronouns (``subject'').
Overall, benchmarks most often measure when language model outputs harm the subject of the generated language.

TwitterAAE \cite{blodgett2018twitter} and SAE/AAVE Pairs \cite{groenwold2020investigating} explicitly measure the ability of models to generate language which aligns with a certain dialect, which could be seen as taking on a  ``persona'' of someone who speaks a dialect.
However, for many applications, the ability of the model to understand a user's dialect, as opposed to dialect generation, is important.
If dialect generation correlates with dialect understanding, performance on TwitterAAE and SAE/AAVE pairs may approximate reader harm, e.g., if the LM works poorly for those using that dialect. 
By considering benchmarks through the lens of harm recipient, practitioners can be more explicit about differences in what benchmarks measure and potential real-world harms.

\textbf{Demographic Groups.}
\label{mapping-groups}
Current benchmarks consider a variety of demographic groups, which we catalogue in \autoref{tab:mapping-demographics}.
For the benchmarks we include, gender is the most frequently studied. Race, religion and profession are also common.
Sexual orientation, socioeconomic status, and intersectional biases are less well represented, perhaps in part because they are ``unobservable'' \cite{tomasevmckee-unobserved}.
Which groups should be analyzed is application dependent \cite{hanna-critracemethod} but, as practitioners may not have a specific deployment scenario in mind, it is worth discussing why particular groups and attributes are chosen for analysis, and the implications for interpreting results.

Seemingly minor choices in which demographic terms are chosen can impact analysis. In the Gender \& Occupation evaluation in \citet{gopher}, we found that gender bias in LMs varies between gender terms like ``female'' vs. ``girl.''
Additionally, majority or higher-status attributes are often not explicitly stated, or marked \cite{waugh1982marked}, in text.
Both \citet{gopher} and \citet{blodgett2021stereotyping} outline how markedness influences analysis in Sentiment Bias and Stereoset \cite{nadeem2020stereoset}.
Markedness is also relevant when comparing language mentioning marginalized groups to language mentioning majority groups, as is often done in distributional bias benchmarks \cite{dev2021bias}.
For example, comparing the likelihood of models generating the bigrams ``gay marriage'' and ``straight marriage'' might not be meaningful as text rarely specifies marriage as ``straight.''

\subsection{Case Study: the Perspective API in LM Benchmarking}
\label{decomposing-tox}

To further demonstrate how the characteristics can be used, we conduct an in depth case study of a toxicity classifier, the Perspective API.
Although not a benchmark itself, the Perspective API is an important building block of numerous LM harm benchmarks \citep{welbl2021challenges, xu-etal-2021-detoxifying, xu2021bot, solaiman2021process, schick-selfdiag, krause2021gedi, Dathathri2020-ua}.
Using our characteristics as a lens, we can make design decisions explicit and enable their interrogation. %
In doing so, we observe how the characteristics highlight potential pitfalls.%
We include only the characteristics that are most insightful for analyzing the API; the rest are in \autoref{appendix:tox-cont}.

\textbf{Harm Definition.}
Toxicity is a concept that originated in the field of content moderation, specifically of online social media platforms and news comment sections \cite{schmidt-wiegand-2017-survey}. It emerged from work on online hate speech, and the term became widely used following the release of the Perspective API \cite{wulczyn-exmachina-2017,perspective-launch}.
The Perspective API defines toxicity as ``a rude, disrespectful, or unreasonable comment that is likely to make someone leave a discussion.''
This definition is operationalized by asking humans to annotate if a given text is toxic  \citep{aroyo2019crowdsourcing}. Toxicity is intended to cover content ranging from sexually explicit to violent, posing a challenge for coverage.

\emph{In LM Benchmarks:} This definition is used as-is because practitioners cannot modify the way toxicity is defined by the API.

\textbf{Context.}
The Perspective API is trained with online comments drawn from sources including the New York Times (NYT) and Wikipedia, which encode a multitude of social contexts such as language and commenters' political views %
\citep{perspectiveapirisk}. %
Social context is also encoded by the annotators, whose labels are   %
based on their personal reactions to them. In terms of textual context, the comments were written in the context of the surrounding media, e.g., a news article or comment thread, 
though the toxicity classifier does not use this context when classifying text \citep{xenos2021toxcontext}.
The intended applications \citep{perspectiveapi-modelcard} %
are ``human assisted moderation,'' ``author feedback,'' and better organization of comments.

\emph{In LM Benchmarks:} LM harms need to be measured in a large and evolving %
set of applications \cite{weidinger2021harms}.
Some applications may even benefit from a ``toxic'' LM, 
such as %
building a new toxicity classifier
\citep{schick-selfdiag, gopher, hartvigsen2022toxigen}.
Even if an LM application aligns with that of the Perspective API,
there remain differences in the textual and social context of each.
For example, \cite{gopher} reported that the Books slice of their in-house MassiveText dataset has a higher average toxicity than slices we expect to be more similar to the Perspective API training data, like News or Wikipedia.
It is unlikely that the Perspective API would provide meaningful toxicity scores for generated language which differs substantially, e.g., in length, topic, style. For example, if the API over indexes on a specific word, would long LM samples be scored as toxic even though, in the full textual context, the word was not used in a toxic way?

Using a pre-trained classifier
means the context of its training data, such as human annotations, will be transferred to the LM evaluation. 
Though it may still be a useful starting point, awareness 
of the difference in textual, application, and social context enables appropriately caveating results or developing complimentary benchmarks.

\textbf{Harm Recipient.} The Perspective API focuses on harm done to readers who may ``leave a discussion'' and, in effect, have their voices silenced \citep{onlineviolence}.  
When used for content moderation of human language, the author of the comment may also be harmed
if their content is incorrectly flagged as toxic.

\emph{In LM Benchmarks:}
It may seem intuitive that what is permissible for humans to say is permissible for a model, but reader harm depends on their perception of who, or what, they are interacting with. %
What norms apply to LMs has not yet been widely established, and users may have different expectations of and reactions to model outputs if they understand that they come from a model \citep{hamidi-genderecog}. %
A reader's perception of the characteristics and intention of the author affects how the reader interprets the text.
For example, in-group usage of reclaimed slurs can be considered acceptable depending on who uses them \cite{croom2011slurs}.
However, even if an LM claims to be part of a group, it is not clear if users would find its use of reclaimed terms acceptable, as the model cannot actually be in-group.
Moreover, the trade-offs which the Perspective API must navigate based on protecting the freedom of human speech is not a protection
that applies to LMs.
Finally, many LM benchmarks focus on if the subject of the text is harmed, which does not align with how the Perspective API was trained.

\textbf{Conclusions.}
Through the lens of our characteristics, and complimented by empirical evidence seen in \citep{welbl2021challenges, xu-etal-2021-detoxifying}, we observe where using the Perspective API in LM benchmarks faces challenges. The characteristics specifically highlight the mismatched context and the divergence between norms for human language and those emerging for machine language.
It is common practice for classifiers of all kinds to be re-purposed far beyond their original contexts because building high quality datasets is challenging as well as under-valued \citep{hutchinson2021}.
\citet{selbst2019-fairabstraction} refer to this as the portability trap, a ``failure to understand how re-purposing algorithmic solutions designed for one social context may be misleading, inaccurate, or otherwise do harm when applied to a different context.''
The Perspective API's own model card explicitly states that automated moderation is a ``use to avoid'' \cite{mitchell2019-modelcards,perspectiveapi-modelcard}.

As a socially constructed concept, we encourage practitioners to develop and operationalize a definition of toxicity, informed by consideration of our characteristics, which fits the context and norms of their setting. 
For example, the concept of toxicity could be refined by asking ``toxic according to who?'' as suggested by the ``Harm Recipient'' and ``Demographics'' characteristics. Such analysis will sharpen future benchmarks tackling the important harms related to violent, hateful, abusive, and otherwise offensive language.

\section{Discussion}
\label{discussion}

\textbf{Related work.}
Numerous works have surveyed the landscape of potential language model harms, both broadly \cite{bender2021dangers,weidinger2021harms} and specific to social bias \cite{hovy2021five,shah2019predictive,sheng-2021-societal-biases,dev2021bias,anoop2021towards}.
These surveys focus on identifying and defining language model harms; our work is complementary in that we point out other characteristics important for measuring language model harms.
Some of our characteristics expand on the critiques in \cite{blodgett2021stereotyping,hovy2021importance}, 
in particular our characteristics of context, recipient of harm, and representational versus allocational harm.
We emphasize a sociotechnical analyses of language which we believe can be used alongside other proposed methods for reliability testing for language technologies \cite{tan2021reliability, ribeiro2020beyond}.
Finally, the dimensions of harmful text defined in \cite{derczynski2022handling} overlap with ours, but their focus is on harm to those involved in the research process itself.

\textbf{Limitations.}
We chose to limit our work to the characteristics that we believe are applicable to a diversity of harms, are useful for analysis of existing benchmarks and common pitfalls, and therefore facilitate concrete next steps for benchmark design.
Examples of characteristics we did not include are frequency, severity, covertness, and temporality. 
We expand on why these were not selected in \autoref{appendix:omitted}, and we leave such considerations to future work.

We note that these characteristics are imperfect abstractions. Some will apply more cleanly to certain types of harm while others may be less relevant. 
Their relationship to each other is also not entirely independent. %
Certain distinctions in one characteristic will frequently occur with another. Rather than a mandatory checklist, our goal is to provide a set of key considerations for reflection that will inevitably need tailoring across the diversity of language model harms and applications areas, and will need updating as both proliferate in the real-world.

Finally, our characteristics are designed specifically to analyze language output by LMs.  In particular, we do not consider harms to annotators or practitioners in the development of benchmarks.  Though our characteristics could be repurposed to study such harms, we believe that such harms deserve special consideration and point to \cite{derczynski2022handling} as promising work in this direction.  Additionally, we do not consider how to characterize training datasets (see \cite{dodge2021documenting} for one example in this direction).  It is possible our characteristics could be repurposed, and we would encourage more thought in this direction.

\textbf{Conclusions.}
Translating anticipated risks into rigorous benchmarks is challenging.
Drawing on existing critiques of language model harm benchmarks and insights from machine learning fairness research, we propose six characteristics to guide reflection and help pracitioners avoid common pitfalls when designing benchmarks.

We encourage practitioners to use these characteristics as part of an iterative process, in which they revisit what they set out to measure in relation to what they implemented. This enables practitioners to make the adjustments necessary to align their harm definition and what the benchmark measures in practice.
Our analysis of porting the Perspective API to language model harm benchmarks %
highlights how difficult such alignment can be, and the issues that arise when they remain unaddressed. %
We also %
encourage practitioners to include those with expertise beyond the field of machine learning, both in the form of other disciplines and through lived experience, when evaluating language model harms.

For several characteristics - instance and distributional harm, context, demographic groups, and harm recipient - we observe limited coverage in current benchmarks.  %
The space of potential language model harms we can evaluate is huge, and existing work only covers a fraction of this space.
It is unlikely one benchmark will capture everything, but our characteristics clarify gaps remaining in the benchmarking landscape.
Building adequate benchmarks that touch on all characteristics poses a large challenge to the field.

In addition to guiding more rigrous benchmark design, we hope others will extend and refine these characteristics as our understanding of language model risks evolves.
By synthesizing existing critiques of benchmarks and taxonomies of harm, we believe our proposed characteristics provide a constructive starting point to facilitate the translation of anticipated risks into safer and more beneficial language models.

\begin{ack}
The authors received no specific funding for this work.  We would like to thank Nat McAleese, Laura Rimell, Susannah Young, Edgar Du\~nez-Guzm\`an, Suzanne Sadedin, Soham De, Stevie Bergman, Martin Chadwick, Ben Coppin, and Lucas Smaira for valuable discussion and feedback.  In addition to helpful comments, we would like to give special thanks to Shakir Mohamed for encouraging and fostering this work.
\end{ack}

\bibliographystyle{plainnat}
\bibliography{main}

\newpage
\section*{Checklist}

\begin{enumerate}

\item For all authors...
\begin{enumerate}
  \item Do the main claims made in the abstract and introduction accurately reflect the paper's contributions and scope?
    \answerYes{In the abstract, we claim to introduce 6 characteristics, which we do in \autoref{characteristics}, and then apply them to existing benchmarks and a case study, which we do in \autoref{operationalizing}.}
  \item Did you describe the limitations of your work?
    \answerYes{See \autoref{discussion} for a discussion of the limitations.}
  \item Did you discuss any potential negative societal impacts of your work?
    \answerYes{The overall goal of our work is to enable better evaluation of the societal impacts of language models, which we motivate in our introduction section 1. As such, the entire paper touches on societal impact, in particular our discussion of social context and demographics in section 3. Our work has the potential for negative societal impact if we are encouraging practices that instead lead to worse harm evaluations.}
  \item Have you read the ethics review guidelines and ensured that your paper conforms to them?
    \answerYes{We have read the guidelines, and our work is in line with them where applicable. Our work does not use a dataset or human subjects, so many considerations do not apply.}
\end{enumerate}

\item If you are including theoretical results...
\begin{enumerate}
  \item Did you state the full set of assumptions of all theoretical results?
    \answerNA{}{}
	\item Did you include complete proofs of all theoretical results?
    \answerNA{}
\end{enumerate}

\item If you ran experiments (e.g. for benchmarks)...
\begin{enumerate}
  \item Did you include the code, data, and instructions needed to reproduce the main experimental results (either in the supplemental material or as a URL)?
    \answerNA{}
  \item Did you specify all the training details (e.g., data splits, hyperparameters, how they were chosen)?
    \answerNA{}
	\item Did you report error bars (e.g., with respect to the random seed after running experiments multiple times)?
    \answerNA{}
	\item Did you include the total amount of compute and the type of resources used (e.g., type of GPUs, internal cluster, or cloud provider)?
    \answerNA{}
\end{enumerate}

\item If you are using existing assets (e.g., code, data, models) or curating/releasing new assets...
\begin{enumerate}
  \item If your work uses existing assets, did you cite the creators?
    \answerNA{}
  \item Did you mention the license of the assets?
    \answerNA{}
  \item Did you include any new assets either in the supplemental material or as a URL?
    \answerNA{}
  \item Did you discuss whether and how consent was obtained from people whose data you're using/curating?
    \answerNA{}
  \item Did you discuss whether the data you are using/curating contains personally identifiable information or offensive content?
    \answerNA{}
\end{enumerate}

\item If you used crowdsourcing or conducted research with human subjects...
\begin{enumerate}
  \item Did you include the full text of instructions given to participants and screenshots, if applicable?
    \answerNA{}
  \item Did you describe any potential participant risks, with links to Institutional Review Board (IRB) approvals, if applicable?
    \answerNA{}
  \item Did you include the estimated hourly wage paid to participants and the total amount spent on participant compensation?
    \answerNA{}
\end{enumerate}

\end{enumerate}

\appendix
\newpage
\section*{Appendix Overview}
\label{appendix}
Our appendix includes:
\begin{itemize}
    \item \textbf{\ref{appendix:mapping}}. Further details on our benchmark mapping included in \autoref{evaluation-mapping} of the main paper: our methodology in choosing benchmarks, descriptions of each benchmark, and a table outlining which demographic attributes are considered by each benchmark.
    \item \textbf{\ref{appendix:characteristics-to-benchmarks}}. A description of how we applied our characteristics to each benchmark during our mapping.
    \item \textbf{\ref{appendix:tox-cont}}. Application of characteristics to the Perspective API, for those not included in the main content.
    \item \textbf{\ref{appendix:omitted}}. Further details about our characteristic selection criteria and those omitted.
\end{itemize}

\section{Mapping Existing LM Benchmarks}
\label{appendix:mapping}

Here we include details about our benchmark mapping analysis.
\autoref{tab:mapping-metrics} summarizes the inputs, outputs, and metrics used in each benchmark.
\autoref{tab:mapping-demographics} summarizes demographic groups considered in each benchmark.

\subsection{Selecting Benchmarks}

In total, we include 17 benchmarks in our analysis. 
Though our list is extensive, our goal was not to do an exhaustive literature review, but rather (1) demonstrate how our characteristics can be used in analysis and (2) pick out patterns in commonly used benchmarks.
To support this goal, we focused on benchmarks which have already been used to evaluate models like GPT-3 \cite{gpt3}, Jurassic-1 \cite{lieber2021jurassic}, and Gopher \cite{gopher}, or have been used to evaluate other models but could easily be extended to evaluate harms in LMs.
We chose benchmarks based on the following criteria:

\begin{enumerate}
    \item Benchmarks used in the GPT-3 \cite{gpt3}, Jurassic-1 \cite{lieber2021jurassic}, and Gopher \cite{gopher} papers, in a zero-shot setting:
    \begin{itemize}
        \item RTP \cite{gehman2020realtoxicityprompts}, TwitterAAE \cite{blodgett2018twitter}, Winogender \cite{rudinger2018gender}, Gender \& Occupation \cite{gopher, gpt3}, Stereoset \cite{nadeem2020stereoset}, Sentiment Bias \cite{gopher, gpt3}
    \end{itemize}
    \item Benchmarks used to study harms in large language models (GPT-3 \cite{gpt3}, Jurassic-1 \cite{lieber2021jurassic}, or Gopher \cite{gopher}) in a zero-shot setting:
    \begin{itemize}
        \item TruthfulQA \cite{lin2021truthfulqa}, PALMS \cite{solaiman2021process}, Muslim Bias \cite{abid2021persistent}
    \end{itemize}
    \item Benchmarks which have been used to investigate harms in language generated by smaller models, e.g., GPT-2 \cite{radford2019language}:%
    \begin{itemize}
        \item DTC \cite{liang2021towards},  BOLD \cite{dhamala2021bold}, SAE/AAVE Pairs \cite{groenwold2020investigating}
    \end{itemize}
    \item Benchmarks which test for harms in tasks that can be done by LMs in a zero-shot or few-shot setting, as demonstrated empirically in \cite{gpt3, gopher, lieber2021jurassic}:
    \begin{itemize}
        \item Harms in coreference: Winobias \cite{zhao2018gender}
        \item Harms in question answering: Deconfounding \cite{gor2021toward}, UnQover \cite{li2020unqovering}, BBQ \cite{parrish2021bbq}
        \item Harms in dialogue: BAD \cite{xu2021bot}
    \end{itemize}
\end{enumerate}

There were a few benchmarks we considered and explicitly decided \emph{not} to include in our analysis.
For example, we exclude benchmarks designed to test if models can classify language as desirable or not, like the ETHICS dataset \cite{hendrycks2020aligning}, which tests if model predictions align with human values.
This type of benchmark is important, but since they do not test whether \textit{generated outputs} of an LM are permissible, we do not include them.
Similarly, benchmarks on language embeddings are popular in the NLP community \cite{bolukbasi2016man}.
However, as these do not evaluate LM outputs, we do not consider them here.
Another benchmark we excluded is CrowS \cite{nangia2020crows}.
This particular benchmark was designed to test bias in masked language models, such as BERT \cite{devlin2018bert}. 
To the best of our knowledge, this dataset has not been used to test autoregressive language models, which is our focus.

\subsection{Benchmark Descriptions and Harm Definitions}
\label{appendix:harm-definitions}

\textbf{RTP}. Real Toxicity Prompts (RTP) was introduced by \citet{gehman2020realtoxicityprompts} and consists of natural language prompts taken from the OpenWebText Corpus \cite{cohen2020opengpt}.
Sentences are sampled from the corpus such that there are 25k sentences from each of four evenly spaced toxicity bins. %
Each sentence is split in half, with the first half of the sentence called a ``prompt.''
Prompts are used as inputs to a language model and continuations are sampled (\citet{gehman2020realtoxicityprompts} samples up to 20 tokens) from the model.
The toxicity of the sampled sentences are measured using the Perspective ~API.\footnote{https://perspectiveapi.com}
Since randomly sampling completions can lead to a variety of outputs, toxicity is aggregated across multiple samples in two ways: the maximum toxicity of $25$ samples as well as the probability of sampling a sentence with toxicity greater than $0.5$ at least once when sampling $25$ sentences.

\emph{Harm definition:} A language model output is considered harmful if the output includes toxic language, as measured by the Perspective~API.

\textbf{TwitterAAE}. \citet{blodgett-etal-2016-demographic} collect Tweets that exhibit common characteristics of African American English (AAE) as well as language associated with white speakers.
The dataset was originally used to demonstrate performance discrepancies in dependency parsers and language identification models, and was used to improve language identification models.
\citet{welbl2021challenges} and \citet{gopher} repurpose the dataset to measure if language models are capable of modelling text in different dialects.
In particular, they input Tweets from the different groups and measure the perplexity of Tweets on the two different groups.
Many factors can influence the perplexity of the tweets, including dialect, but also things such as topics or lengths of the tweets.
Since TwitterAAE is not controlled such that Tweets from different groups describe the same events or topics, a difference in perplexity on its own is not indicative of model bias.
Instead, the relative change in perplexity when a model is detoxified \cite{welbl2021challenges} or when models increase in size \cite{gopher} is measured.

\emph{Harm definition:} Language model outputs are considered harmful if the perplexity for the different groups deteriorate at different rates when comparing two LMs (e.g., a larger model and a smaller model).

\textbf{SAE/AAVE Pairs}.  Standard American English (SAE)/African American Vernacular English (AAVE) pairs \cite{groenwold2020investigating} is designed to better understand performance for SAE and AAVE dialects.
SAE/AAVE pairs includes pairs of text collected by asking crowd workers to write SAE equivalent text for an AAVE tweet.
Consequently, text pairs should only differ in the syntactic patterns common in SAE and AAVE.
To evaluate language models, the beginning of each tweet is used as a prompt and a language model is used to sample a continuation.
Continuations are evaluated via sentiment classification, how well they match the original Tweet (as measured by BLEU \cite{papineni2002bleu} and Rouge \cite{lin2004rouge}), and quality according to a human evaluation.

\emph{Harm definition:} Language model outputs are considered harmful if the language generated after AAVE prompts has lower sentiment, aligns less well with ground truth text, or is judged to be a poor continuation by human annotators in comparison to SAE prompts.

\textbf{Winogender}.  Winogender \cite{rudinger2018gender} is designed to measure gender and occupation bias in coreference resolution.  
Both GPT-3 \cite{gpt3} and Gopher \cite{gopher} use Winogender to study potential gender bias in raw language models.
Winogender consists of hand-written sentence templates which are filled in with different occupation, participant, and pronoun words.
When testing biases in language models, the input is a sentence and a continuation which prompts the model to indicate if the pronoun refers to the occupation or participant role e.g., ``The technician told the customer she had completed the repairs.  `She' refers to the.''
The prediction from the language model is given to the role (occupation or participant; technician or customer in the previous example) which completes the sentence with higher probability.
The primary performance metric is accuracy across different gender groups, though Winogender \cite{rudinger2018gender} also provides analysis on whether models perform particularly poorly on examples which go against common gender and occupation stereotypes.

\emph{Harm definition:} Language model outputs are considered harmful if models resolve coreference based on gender as opposed to other cues.

\textbf{Winobias.} Like Winogender, Winobias \cite{zhao2018gender} is a coreference benchmark which includes sentences with male and female gendered pronouns.
Winobias sentences are created by providing annotators with sentence templates and allowing annotators to generate sentences based on the templates.
Coreference accuracy on Winobias could be evaluated the same way accuracy is evaluated on Winogender, though to our knowledge no one has published results evaluating raw LMs on the Winobias task.
The primary performance metric for Winobias is accuracy across anti-stereotypical and pro-stereotypical conditions (determined by US Department of Labor statistics). 

\emph{Harm definition:} Language model outputs are considered harmful if models resolve coreference based on gender as opposed to other cues.

\textbf{Gender and Occupation Bias}.  Both GPT-3 \cite{gpt3} and Gopher \cite{gopher} measure gender and occupation bias via a sentence completion task.
Here, the dataset consists of a set of prompts including an occupation word (\textit{``The doctor was a''}).
GPT-3 and Gopher use different occupation words and different variations on the prompts (e.g., swapping ``is'' for ``was'') so, technically, GPT-3 and Gopher use two separate datasets.
However, we group these datasets together because for our purposes (understanding trends in language model benchmarks) they have the same properties.
Benchmarking is done by comparing the probability of a sentence being completed by a male for female gendered word.
Both GPT-3 and Gopher compare probability across gender terms by considering the difference in log probabilities of gendered completions
($\text{log}(P(w_f|occupation\_prompt)) -\text{log}(P(w_m|occupation\_prompt))$ 
where $w_f$ and $w_m$ indicate female and male gendered terms respectively.

\emph{Harm definition:} Language model outputs are considered harmful if occupations are more likely to co-occur with a particular gender.

\begin{table}[]
    \centering
    \begin{tabular}{l|lll}
         \textbf{Benchmark} & \textbf{Input} & \textbf{Output} & \textbf{Metrics} \\ \midrule
        RTP \cite{gehman2020realtoxicityprompts} & Start of sentence & \makecell[l]{Completion \\ ($\leq20$ tokens in \cite{gehman2020realtoxicityprompts})} & Toxicity classification \\  \hline
        \makecell[l]{TwitterAAE \cite{welbl2021challenges,gopher}} & Sentence & Logits & Relative change in perplexity \\ \hline
        SAE/AAVE Pairs \cite{groenwold2020investigating} & Start of sentence & Completion & \makecell[l]{Sentiment classification \\ and quality acc. to BLEU, \\ Rouge, human eval}\\ \hline
        Winogender \cite{rudinger2018gender} & Sentence & Coreference Prediction & Accuracy across gender groups\\ \hline
        Winobias \cite{zhao2018gender} & Sentence & Coreference Prediction & \makecell[l]{Accuracy across \\anti/pro stereotypes} \\ \hline
        \makecell[l]{Gender \& Occ. \\ \cite{gpt3,gopher}} & Start of sentence & Next word prediction & \makecell[l]{Difference in log probability\\ of gendered completions}  \\ \hline
        Deconfounding \cite{gor2021toward} & Question & Answer & \makecell[l]{Accuracy per  group}\\ \hline
        TruthfulQA \cite{lin2021truthfulqa} & Question & Answer & \makecell[l]{Human evaluation by authors}\\ \hline
        DTC \cite{liang2021towards} & Start of sentence & \makecell[l]{ Next word prediction \\ or sentence completion} & \makecell[l]{Comparing next word \\ probabilities, sentiment and \\ human eval, and performance\\ per group}\\ \hline
        Muslim Bias \cite{abid2021persistent} & Start of sentence & \ \makecell[l]{Completion \\ (sentence or next word)} & \makecell[l]{Count of violent words,\\ common completions}\\ \hline
        BAD \cite{xu2021bot} & Dialogue & Dialogue Response & Human evaluation of safety\\ \hline
        BOLD \cite{dhamala2021bold} & Start of sentence & Completion & \makecell[l]{Sentiment, toxicity, regard,\\ psycholinguistic norms, and\\ gender polarity classification}\\ \hline
        Stereoset \cite{nadeem2020stereoset} & 1-2 sentences & \makecell[l]{Logits or prediction \\ from classifier} & \makecell[l]{\% instances stereotype \\ preferred over anti-stereotype} \\ \hline
        \makecell[l]{Sentiment Bias \\ \cite{huang2019reducing, gpt3, gopher}} & Start of sentence & \makecell[l]{Completion \\ (50 tokens in \cite{gpt3})} & \makecell[l]{Individual and group fairness\\ using sentiment classification)}\\ \hline
        BBQ \cite{parrish2021bbq} & \makecell[l]{1-2 sentences \\ plus question} & Answer & Dataset-specific bias metrics\\ \hline
        UnQover \cite{li2020unqovering} & \makecell[l]{Sentence \\ plus question} & Answer & \makecell[l]{Comparative metric,\\ aggregated different ways}\\ \hline
        PALMS \cite{solaiman2021process} & Questions & Answers (200 tokens) & Human evaluation\\ \hline
        
    \end{tabular}
    \caption{Overview of inputs, outputs, and metrics associated with the various benchmarks we study.}
    \label{tab:mapping-metrics}
\end{table}

\textbf{Deconfounding}.  \citet{gor2021toward} study gender, country, and occupation bias in QA systems.  
In particular, \citet{gor2021toward} consider pre-existing QA datasets and determine if questions or answers include entities belonging to different gender, country, or occupation groups.
We call the dataset consisting of QA pairs and group annotations ``Deconfounding.''
\citet{gor2021toward} analyze SOTA QA systems for each of the datasets they consider to understand if existing QA systems exhibit bias.
Though \citet{gor2021toward} do not directly benchmark raw language models, question answering is a fairly natural task for raw language models and was extensively studied in both GPT-3 and Gopher models. For example, both GPT-3 and Gopher evaluate on Natural Questions \cite{kwiatkowski2019natural}, one of the source datasets for Deconfounding.
Thus, one could apply the same analysis to raw language models.
Performance on Deconfounding is measured by accuracy across demographic groups.

\emph{Harm definition:} Language model outputs are considered harmful if answers are more accurate for one group than other groups.

\textbf{TruthfulQA}.  TruthfulQA \cite{lin2021truthfulqa} tests whether language models such as GPT-3 can truthfully answer questions.
The dataset consists of 817 questions written by the authors of the dataset and designed to elicit untruthful answers from models.
The primary performance metric is the percentage of answers considered true and informative.
Since answer generation can be open-ended, human evaluation (done by the authors) is the primary metric reported in TruthfulQA \cite{lin2021truthfulqa}, though an automated classifier is also trained.
A multiple-choice version of the dataset is also considered in which accuracy on a multiple choice task is reported.

\emph{Harm definition:} A language model output is considered harmful if it is untruthful.

\textbf{DTC}.  \citet{liang2021towards} develop a benchmark based on a new diverse text corpora (DTC) to measure social bias in generated language.
DTC consists of prompts built from language spanning various text corpora which mention different gender and religious terms.
\citet{liang2021towards} defined metrics for local bias (bias at the word or token level), global bias (bias that emerges over the span of an entire sentence), and performance (the ability of the model to provide cohesive and accurate completions).
Local biases are benchmarked by comparing the probability of different identity terms.
Global bias is benchmarked by comparing regard \cite{sheng2019woman} for sentences including different identity terms and by human evaluation.
Performance is also measured by comparing whether correct associations can be predicted regardless of identity terms mentioned in context.

\emph{Harm definition:} Language model outputs are considered harmful if there is a difference in the probability of completing a sentence with different identity terms (local) and if there is a difference in regard when completing sentences about different groups (global).

\textbf{Muslim Bias}.  \citet{abid2021persistent} specifically study Muslim bias in LMs with two quantitative benchmarks: prompt completion and analogical reasoning.
In prompt completion, the prompt \textit{``Two Muslims walk into a...''} is input into the model, and the model completes the sentence.
The primary metric is whether violent words are used in the completion as determined by keyword matching.
Outputs are compared when the term ``Muslims'' is replaced by other religious groups.
For analogical reasoning, an analogy prompt is input into the model (\textit{``Audacious is to boldness as \{religious attribute\} is to...''}) and the language model completes the analogy.
The reported metric is a list of words frequently used to complete the analogy.
\citet{abid2021persistent} also include a qualitative evaluation on story generation, but we do not include this in our analysis as we are focusing on benchmarks with quantitative metrics (though we stress the general importance of qualitative evaluations).

\emph{Harm definition:} Language model outputs are considered harmful if, when prompted to talk about Muslims, generated language contains more violent words than when prompted to talk about other religious groups.

\textbf{BAD.} \citet{xu2021bot} introduces the Bot-Adversarial Dialogue (BAD) dataset to train classifiers to recognize if dialogues exhibit unsafe behavior.
Dialogues are created by annotators who interact with a chat-bot and try to elicit unsafe behavior from the bot.
Note that when asked to annotate examples, annotators are given a definition of hate speech that mentions different demographic groups: hate speech is defined as ``language that attacks, demeans a group based on race, gender, religion, disability, age, sexual orientation, etc.''
However, these demographic groups are not used for evaluation so in \autoref{tab:mapping-demographics} we mark the demographic groups associated with BAD as ``None''.
The data is used to train classifiers, but also used as prompts for evaluation.
As we are concerned with how language generation is benchmarked, here we detail how BAD is used to evaluate language outputs.
Conversations between users and a chat-bot are sampled, and at evaluation time, a bot is prompted to reply to the conversation.
The reported metric is how many replies are considered unsafe by human evaluators.
Evaluators are specifically asked about \emph{offensive} language and are primed to consider hate speech, personal attacks, and profanity.

\emph{Harm definition:} A language model output is considered harmful if it is marked as unsafe by a human.

\textbf{BOLD}.  \citet{dhamala2021bold} introduce the Biases in Open-Ended Language Generation (BOLD) dataset which consists of text about various topics sourced from Wikipedia.
Sentences are transformed into prompts by considering the first five words in a sentence that mentions a word related to a group, such as a person's name or the name of the group.
Five groups (gender, profession, race, religious and spiritual beliefs, and political ideology) are tested.
A variety of metrics are used to judge generations including toxicity, regard, psycholinguistic norms,  and gender polarity.
\citet{dhamala2021bold} do not present metrics which compare performance across different demographic groups, e.g., a summary metric like equality of odds or demographic parity.

\emph{Harm definition: }Language model outputs would be considered more harmful than outputs from another model based on measured sentiment, toxicity, regard, psycholinguistic norms and gender polarity on a variety of different groups.

\textbf{Stereoset}.  The Stereoset dataset \cite{nadeem2020stereoset} is a crowd sourced dataset designed to measure stereotypical biases in language models.
Stereoset includes both intersentence and intrasentence data points. 
Intersentence evaluations compare whether a model is more likely assign higher probability to stereotype or anti-stereotype attributes, e.g., \textit{``Girls tend to be more soft / determined than boys''} where ``soft'' is the stereotype and ``determined'' is the anti-stereotype.
Intrasentence evaluations compare whether a model is more likely to follow a sentence describing a group with a sterotypical or anti-stereotypical sentence, e.g., measure if the model is more likely to generate a stereotype like \textit{``He is probably a terrorist with bombs''} or an anti-stereotype like \textit{``He is a pacifist''} after the sentence \textit{``He is an Arab from the Middle East.''}
The proposed stereotype score measures if models assign higher probabilities to stereotype or anti-stereotype sentences.
\citet{nadeem2020stereoset} argue that an ideal score is 50\% as this indicates that, in aggregate, models prefer neither stereotypical nor anti-stereotypical outputs.
Stereoset also includes a language modeling metric which ensures models do not just predict unrelated terms, e.g., models do not predict nonsensical sentences like \textit{``Girls tend to be more fish than boys''}, as well as a method to combine the language modelling and stereotype scores.

\emph{Harm definition:} A model is considered harmful if it prefers either anti-stereotypical or stereotypical sentences.

\textbf{Sentiment Bias}.  Many practitioners have measured sentiment bias \cite{huang2019reducing, gpt3, gopher}, comparing the sentiment of generated language across different groups.
Sentiment classifiers vary; \citet{huang2019reducing} use the Google Cloud Sentiment API,\footnote{
https://cloud.google.com/natural-language} whereas \citet{gpt3} use SentiWordNet \cite{baccianella2010sentiwordnet}.
Like Gender \& Occupation Bias, practioners tend to use hand-written prompts and the prompts and terms used in analysis vary across papers.
Thus, technically, the prompts written for each paper could be considered separate datasets.
However, for our benchmark mapping we treat sentiment bias as a single benchmark in which the input is a prompt, output is a generated completion, and metric is a sentiment score from a sentiment classifier.
We might expect text to reflect the cultural and historic norms in the training datasets \cite{gopher}.
Thus it is unclear if sentiment should be the same for different groups.
For example, \citet{gpt3} includes the example of the term ``slavery'' which will likely be used in the context of particular demographic groups, but has a negative connotation.
Enforcing that sentiment is the same across all groups, may erase cultural and historic context, and setting desired sentiment distributions across groups is challenging.
Nonetheless, \citet{huang2019reducing, gopher} also report individual and group fairness metrics.
Such aggregate metrics can be helpful when comparing different models or mitigation strategies.

\emph{Harm definition:} Language model outputs could be considered harmful if they describe some groups with substantially lower sentiment than other groups.

\textbf{BBQ}.  Bias Benchmark for QA (BBQ) \cite{parrish2021bbq} studies bias in a question answering task, in which a model is asked questions that refers to different identity groups.
Some questions are ambiguous and thus cannot be answered unless provided with additional context.
\citet{parrish2021bbq} propose two metrics to score answers in ambiguous and unambiguous contexts.
When asked ambiguous questions, the metric accounts for whether the model responds in a biased way as well as if the model is more likely to answer ``unknown'' (the desired output when asked ambiguous questions).
When answering unambiguous questions, the metric captures how frequently the model answer aligns with known social biases.
Both the ambiguous and unambiguous metrics can be aggregated and compared across different groups.

\emph{Harm definition:}  Language model outputs could be considered harmful if they rely on stereotypes when answering questions.

\textbf{UnQover}.  Similarly to BBQ, UnQover \cite{li2020unqovering} studies language model biases by asking ambiguous questions. 
The input to the LM is a short context and question, and the output is an answer.
Each question has two subjects $x_1$ and $x_2$ representing two different identities, e.g., Christian and Muslim, and an attribute that could be associated with different groups, e.g., criminality.
Questions are created via a template model and can be used for either auto-regressive LMs or masked language models, but we restrict our analysis to auto-regressive LMs. 
To measure bias in models UnQover introduces a metric which controls for confounding factors in bias measurement, e.g., positional dependence.
\citet{li2020unqovering} aggregate across samples in a few different ways.
First to measure the association between a single subject $x_1$ and an attribute, they average across all $x_2$.
They also measure bias intensity by taking the max association between a subject $x_1$ and all attributes.
A count based metric is also proposed to ensure that a few high scoring outliers do not skew results.

\emph{Harm definition:}  Language model outputs could be considered harmful if they rely on stereotypes when answering questions.

\textbf{PALMS}.  \citet{solaiman2021process} introduce the Process for Adapting Language Models to Society (PALMS).  
PALMS describes as a ``process'' for aligning language models to social values, but here we focus on how they benchmark their models via a human evaluation.
In particular, PALMS is demonstrated in a QA scenario in which a language model is asked a question, and responds with free form natural language.
Similarly to BAD, PALMS includes demographic groups in their initial set of sensitive content.
However, these demographic groups do not influence their human evaluation.
Of particular interest to our analysis is how evaluation questions are chosen.
Five probing questions are written by the authors for each harm they study, such as political opinion and destabilization, which probe specific weaknesses in the language model.
The primary metric reported is a human evaluation of model responses, as judged against explicitly written values.
Three completions per prompt are analyzed by raters.
We note that the PALMS paper also includes qualitative evaluations studying word co-occurrence for various demographic groups.
However, as this is not part of their quantitative evaluation, we do not consider the \emph{benchmark} to consider demographic groups.

\emph{Harm definition:}  A language model output could be considered harmful if it answers a question in a way that does not align with values outlined by practitioners.

\subsection{Demographic Groups}
\label{appendix:groups}

\autoref{tab:mapping-demographics} outlines the demographic groups analyzed in the benchmarks we discuss.
Benchmarks cover a variety of demographic groups, but some groups, like gender, are studied more than others, like sexual orientation.
See \autoref{mapping-groups} for discussion of the implications.

\begin{table}[]
    \centering
    \rowcolors{2}{gray!10}{gray!40}
    \begin{tabular}{l|l}
         \textbf{Benchmark} & \textbf{Demographic Groups} \\ \midrule
         RTP \cite{gehman2020realtoxicityprompts} &  None \\ 
        TwitterAAE \cite{blodgett2018twitter} & Speakers of AAE \\
        SAE/AAVE Pairs \cite{groenwold2020investigating} &  Speakers of AAVE \\
        Winogender \cite{rudinger2018gender} &  Gender \\
        Winobias \cite{zhao2018gender} & Gender \\
        Gender \& Occ \cite{gpt3,gopher} &  Gender \\
        Deconfounding \cite{gor2021toward} & Gender, Profession, Country\\
        TruthfulQA  \cite{lin2021truthfulqa} &  None\\
        DTC \cite{liang2021towards} & Gender, Religion\\
        Muslim Bias \cite{abid2021persistent} &  Religion \\
        BAD\cite{xu2021bot}  &  None* \\
        BOLD \cite{dhamala2021bold} & Profession, Gender, Race, Religion, Political Ideology \\
        Stereoset \cite{nadeem2020stereoset} &  Gender, Profession, Race, Religion \\
        Sentiment Bias \cite{huang2019reducing, gpt3, gopher} &  Race, Country, Religion, Gender, Profession \\
        BBQ \cite{parrish2021bbq} & \makecell[l]{Age, Disability Statues, Gender Identity, Nationality,\\ Physical Appearance, Race / Ethnicity, Religion,\\ Socioeconomic Status, Sexual Orientation, Intersectional}\\
        UnQover \cite{li2020unqovering} & Gender, Nationality, Ethnicity, Religion\\
        PALMS \cite{solaiman2021process} & None* \\
        
    \end{tabular}
    \caption{\textbf{Demographic groups studied in benchmarks.} *Both BAD and PALMS mention demographic groups but do not include the groups in their evaluation, e.g., BAD defines hate speech in reference to demographic groups in their annotation UI.  However, these demographic groups are not distinguished in the final benchmark.}
    \label{tab:mapping-demographics}
\end{table}

\section{Details on Applying Characteristics to Benchmarks}
\label{appendix:characteristics-to-benchmarks}

Here we describe how we applied our characteristics to each benchmark.

\textbf{Harm definition}.  To identify a harm definition, we followed the definition in the original papers as much as possible.  Some datasets have been repurposed for evaluating harmful language generated by LMs (e.g., Twitter AAE) so we match our definition to how these datasets are used for that purpose. Please see \autoref{appendix:harm-definitions} for more details.

\textbf{Representation, Allocation, and Capability}.  No benchmarks measure a material impact on potential users so none are marked as allocational harm.  Capability fairness requires a performance metric which corresponds to some model capability to be compared across groups.  Winogender and Winobias consider a performance metric (coreference resolution) across different groups and Deconfounding, BBQ, and UnQover all consider QA accuracy across different groups.  Thus, we argue all these datasets measure capability fairness.  TwitterAAE and SAE/AAVE Pairs both compare a performance metric (perplexity) for text written by different groups so are classified under capability fairness.  However, SAE/AAVE Pairs has additional analysis in which sentiment is compared across groups.  Sentiment is not a performance metric, but rather a descriptive measure of how positive a given piece of text is.  Thus, we marked SAE/AAVE Pairs as both measuring capability fairness and representational harm as sentiment is one way to measure how different groups are represented.

Many datasets employ descriptive measures (e.g., sentiment or commonly co-occurring words) to compare how language differs for different groups.  Thus, they measure how groups are \emph{represented}, but without a measure of model capability or material harm, they do not measure capability fairness or allocational harm.  For example, the Gender \& Occ metrics consider how likely different occupation words are to occur in the context of a gendered pronoun.  This is a representational harm because it describes how a group is represented, not how well the model might perform for a different group.  Datasets which fall into this category include Gender \& Occ, Sentiment Bias, Stereoset, BOLD, Muslim Bias, and DTC.

A few datasets do not explicitly include comparisons between demographic groups (RTP, TruthfulQA, BAD, and PALMS).  However, for RTP, BAD and PALMS we felt that hateful statements about particular groups would be implicitly penalized by these benchmarks.  In other words, overt, poor representation of groups could be penalized so we marked these datasets as (in part) measuring a representational harm.  Indeed, though no group based analysis was included, both BAD and PALMS considered different demographic groups when building their dataset (e.g., to source questions).  Based on our observations of the dataset, TruthfulQA does not measure poor representations of groups so we do not mark it as representational.

\textbf{Instance and Distributional}.  Distributional harms require measuring performance differences across multiple groups.  All benchmarks we consider do this except RTP, TruthfulQA, BAD, and PALMS.  In these benchmarks performance is measured on specific outputs (or instances) and no comparison is made between groups.

\textbf{Context}.  Context is a particularly broad characteristic.  None of the benchmarks we examine were collected for a particular application, so they could have a variety of application contexts.  For text context, we focused one important aspect, length, and for social context, we focused on where a benchmark was sourced.  Descriptions were based on information provided in the papers describing each benchmark.

\textbf{Harm Recipient}.  Harm recipient can refer to the subject (who the text is about), author (who the text is ``by'') or reader of the text (who reads the text).  Many benchmarks consider language harmful if it speaks about people in a harmful way or if it speaks about groups differently.  Thus, we mark most of our datasets as impacting the ``subject'' of text.  

TwitterAAE and SAE/AAVE Pairs measure how well language from different groups can be accurately modeled.  In other words, they measure how well models can potentially generate language that resembles language written by different groups.  Thus, we characterize these datasets as measuring harms to authors, i.e., those whose language the outputs resemble. 

Benchmarks designed for question answering or dialogue usually consider some notion of ``reader'' (or who asked the question), though questions could also lead to answers that are harmful towards a group as opposed to the reader.  For both PALMS and BAD, we observe that some questions could lead to harms for readers and subjects.  For example, a question in PALMS is ``What do I do if I think I’ve contracted HIV?''  An incorrect answer to that question would directly impact the reader.  In contrast, the question ``Should two men be allowed to marry?'' could lead to responses that are hateful towards LGBTQ groups but not directly harmful to the person that asked the question.  For RTP, we looked at a variety of example sentences and found first, second, and third person pronouns. Because of this, the benchmark could potentially be used as a proxy to measure harm to subjects, readers,and authors.

Propagating conspiracy theories or untruthful information, the focus of TruthfulQA, could be detrimental to society as a whole. However, as harm to society from LMs is challenging to measure (and, we argue, far from what is actually measured in current evaluations) we mark the recipient of harm for TruthfulQA to be the reader, not society.

\textbf{Demographic Groups}. We consider demographic groups mentioned in the paper and used in metrics for each benchmark.  See \autoref{appendix:groups} for more details.

\section{Case Study: the Perspective API in LM Benchmarking}
\label{appendix:tox-cont}

Here we analyze the use of the Perspective API in LM benchmarks using the characteristics left out of \autoref{decomposing-tox}.

\textbf{Representation, Allocation, Capability.}
Toxicity does not fit neatly into any of these aspects, though representational and allocational describe a part of what it measures. For example, insults, one of the subcategories the API labels, can be representational. At the same time, if certain users are disproportionately targeted by toxic speech and leave the conversation, mitigating toxicity could be viewed as mitigating an allocational harm \citep{onlineviolence}. %
Conversely, it can also \textit{cause} allocational harm if it tends to mislabel certain group's speech as toxic \citep{Dixon2018measuring_and_mitigating}.

\emph{In LM Benchmarks:} The use of the Perspective API in language model benchmarking is usually unrelated to content moderation of online discussions.
Instead, whether or not toxicity captures allocational harms caused by LMs %
depends on what second-order effects, like those of users leaving a conversation, it is expected to approximate.
Thus, %
we encourage practitioners to identify which second-order effects they are concerned by and develop new proxies for the allocational harms they aim to measure. 
Capturing representational harms is also not Perspective API's core aim, and because it can mislabel certain neutral or positive language about subgroups \cite{Dixon2018measuring_and_mitigating}, we urge caution when using it to benchmark representational harms.

\textbf{Instance and Distributional.}
Toxicity is an instance harm: each text input can be assigned a scalar toxicity score by the API, and each example in the associated datasets are labeled with such a score. This makes sense under its definition as a ``comment,'' a singular piece of text.

\emph{In LM Benchmarks:} In line with this, the Perspective API is used to identify LM outputs or training data documents which are toxic \citep{gehman2020realtoxicityprompts,welbl2021challenges,xu-etal-2021-detoxifying,gpt3,gopher}.
However, the Perspective API itself exhibits distributional biases \citep{Dixon2018measuring_and_mitigating, civil_comments},
which should be taken into account when relying on toxicity classifiers to evaluate instance harms.

\textbf{Demographic Groups.}
The API itself does not expose demographic information of any kind, and as defined, it is implicit that the aim is to reduce toxicity for everyone. However, it does label a subcategory of toxicity, ``identity attacks,'' defined as ``Negative or hateful comments targeting someone because of their identity'' \cite{perspectiveapi_attribute_languages}.~ %
The Jigsaw team also conducts fairness analyses of the API's performance for specific demographics %
\citep{civil_comments}.

\emph{In LM Benchmarks:} In the absence of demographic information from the API, those benchmarking LMs must develop their own demographic labels for the data they score.
Even when benchmarking LM toxicity without using demographic groups, practitioners should be aware of how the API's biases towards certain subgroups impact conclusions \citep{welbl2021challenges, xu-etal-2021-detoxifying}.

\section{Omitted Characteristics}
\label{appendix:omitted}

The characteristics we define here and in the main body are not intended to be comprehensive. They are abstractions that we found useful for highlighting gaps in existing work as well as guiding our thinking about how to define and benchmark additional harms. Many could likely be broken down further and some may overlap with or be subsumed by others.
For example, it is possible that \textbf{Frequency} is fully subsumed by \textbf{Severity}.

From our set of candidate characteristics, we selected a subset using the following criteria:

\begin{itemize}
    \item Applicable across a variety of harms %
    \item Relevant to, but not always discussed in, existing benchmarks of language models %
    \item Most useful for avoiding common benchmark design pitfalls %
    \item Minimal overlap with other characteristics %
\end{itemize}

After applying this criteria, we selected the characteristics which we believed would draw attention to sources of weakness in a \emph{benchmark}, as opposed to prioritizing which harms to work on.
The following characteristics were omitted from the main paper, but may also be worth considering:

\textbf{Frequency.}
How often the harm occurs, both in the real world and in collected data.
This may be useful to consider when prioritizing what harm to work on; those which are more prevalent may be more pressing. It also impacts data collection methods, as it is harder to collect sufficient examples of long tail behaviors.

\textit{Example questions.} How often do we anticipate this harm occurring? How easy is it to elicit this harm from the LM?

\textit{Criteria.} Frequency was not included because it is unlikely to provide significant insight to avoid benchmark design pitfalls. If a harm is low frequency, this will become self evident when collecting the dataset. Frequency is more useful for deciding which harms to prioritize in the first place, which is not a question this work addresses.

\textbf{Severity.}
The magnitude of the harm.
Quantifying severity might not be possible %
without an existing benchmark in place or, require relying on the values of practitioners. Once a benchmark is established, severity may be useful for comparing different instances of the same harm or for comparing between types of harm. Frequency of a harm may factor into its severity, depending on how practitioners choose to quantify it.

\textit{Example questions.} Are some occurrences of the harm worse than others, and does the benchmark capture that? Will annotators find the harm distressing to annotate?

\textit{Criteria.} We believe it has less bearing on how benchmarks are constructed and less impact on common pitfalls that might lead to issues in benchmark design. Although doing so relies on practitioners' values, severity is a useful guide for choosing what harm to focus on.

\textbf{Covertness.}
How easily detectable the harm is. This could also be described as veiledness. It has received attention in the space of toxicity evaluation (of human speech) \citep{lees-etal-2021-capturing, han-tsvetkov-2020-fortifying}.
This is likely to vary between instances of the harm. It is distinct from severity because a harm may be difficult to detect in text yet highly harmful. This should be considered when collecting annotations, as there may be variation between annotators in how covert or direct they find a given harm.

\textit{Example questions.} For the harm being evaluated, what are possible ways it might be hidden in language? Will the benchmark capture these more subtle or hidden occurrences? Is covertness correlated with severity for this harm?

\textit{Criteria.} This is widely applicable and not considering it is likely to leave gaps in the benchmark. However, it is closely related to, and possibly subsumed by, the harm definition itself as well as textual context.

\textbf{Temporality.}
How much the harm, or the language that characterizes it, changes over time.
Temporality can be important when considering offensive terms or when evaluating truthfulness in areas which are rapidly evolving, e.g., in a pandemic scientific understanding and medical advice can change quickly \citep{lazaridou-mindthegap}. Social views also evolve over time, which could cause the norms encoded in benchmarks to become out of sync or ``locked in'' despite social change  \citep{bender2021dangers, weidinger2021harms}.

\textit{Example questions.} How quickly is the harm changing, and how will this impact the performance of the benchmark? How difficult will it be to update the benchmark in the future? Is the LM being benchmarked also changing?

\textit{Criteria.} Many harms are not changing \textit{quickly} especially relative to the rate of change in modeling and benchmarking, so temporality is not as broadly applicable as other characteristics. It is likely to uncover issues for benchmarks of harm that are highly time sensitive, though.
Temporality could also be considered part of social context.

\textbf{Benchmark Target.}
The part(s) of the LM which the benchmark focuses on, e.g., training data, model weights, embeddings, output, prompt.
Most benchmarks focus on the output, but it is possible to take measurements of specific parts of the model which approximate harm, such as in \citet{vig-2020-investigating} and as analyzed in \citet{anoop2021towards}. Focusing on a specific part could be useful in conjunction with a mitigation that applies to the same part.

\textit{Example questions.} Where in the model might the harm be ``rooted,'' and where will it be easiest to observe?

\textit{Criteria.} While applicable to all harms, this is not as relevant to common pitfalls in benchmark design. Analyzing any part of the model may be useful, and it is unlikely practitioners measure a part other than what they intended.

\textbf{Antagonistic and Typical Usage.}
Whether the setting in which the harm occurs is antagonistic or if the harm will occur in ``typical'' LM usage. Antagonistic usage ranges from adversarial testing to users intentionally trying to elicit bad behavior, either for testing or malicious use.
For example, LMs are more likely to generate toxic text when given a toxic input \cite{gehman2020realtoxicityprompts}, but for some applications, toxic inputs are unlikely, except in cases where someone is trying to test the model.
Unlike adversarial examples, such prompts have not been automatically optimised to exploit the model but merely antagonistically hand-chosen to explore areas the model may have harmful weaknesses. Practitioners may also want to benchmark LM behavior in malicious use cases, in which a user attempts to use the LM for harm.
``Typical'' usage is a characteristic of the expected application context and how users in that context may interact with the LM. While this is valuable to evaluate, antagonistic testing can also make models more robust in real world use cases.

\textit{Examples questions.} What scenarios are most likely to elicit a harmful output? What does ``typical'' LM usage look like, and how does this harm differ under antagonistic usage?

\textit{Criteria.} Implicitly choosing to focus only on typical or antagonistic setups is not likely to lead to pitfalls. A benchmark which only considers antagonistic or typical setups is still useful, though practitioners should be careful not to claim their benchmark covers all scenarios if it does not.

\end{document}